\documentclass[sigconf,natbib=true,review=false, anonymous=false]{acmart}

\usepackage{acronym}
\usepackage{booktabs} % For formal tables
\usepackage[skip=0pt]{caption}
\usepackage{multirow}
\usepackage{bm}
\usepackage{bbm}
\usepackage{balance}
\usepackage{lineno}
\usepackage{subfig}
\usepackage[inline]{enumitem}
\usepackage{pifont}
\usepackage{array}
\usepackage{graphicx}
\usepackage[ruled,commentsnumbered,linesnumbered]{algorithm2e}
\usepackage[T1]{fontenc}
\usepackage{aecompl}
\usepackage{color}
\usepackage{threeparttable}

\usepackage{url}
\usepackage[T1]{fontenc}

\acrodef{PLM}{pre-trained language model}
\acrodef{ourMethod}{Our Method Name}
\acrodef{EHR}{electronic health record}
\acrodef{ICD}{international classification of diseases}
\acrodef{NLP}{natural language processing}
\acrodef{MSE}{mean square error}
\acrodef{PMI}{point-wise mutual information}
\acrodef{MDG}{attention map discrimination guiding}
\acrodef{PDG}{attention pattern decorrelation guiding}
\acrodef{AG}{attention guiding}
\acrodef{PCA}{Principal Component Analysis}

\newcommand{\header}[1]{\vspace*{1mm}\noindent{\textbf{#1}}}

\setcopyright{rightsretained}
\copyrightyear{2022}
\acmYear{2022}
\acmDOI{10.1145/1122445.1122456}

\acmPrice{15.00}
\acmISBN{978-1-4503-XXXX-X/18/06}


\author{Shanshan Wang}
\affiliation{%
  \institution{Shandong University}
  \city{Qingdao}
  \country{China}
}
\email{wangshanshan5678@gmail.com}

\author{Zhumin Chen $^*$}
\affiliation{%
  \institution{Shandong University}
  \city{Qingdao}
  \country{China}
}
\email{chenzhumin@sdu.edu.cn}

\author{Zhaochun Ren}
  \affiliation{%
  \institution{Shandong University}
  \city{Qingdao}
  \country{China}
}
\email{zhaochun.ren@sdu.edu.cn}

\author{Huasheng Liang}
 \affiliation{
 \institution{WeChat, Tencent}
 \city{Guangzhou}
 \country{China}
}
\email{watsonliang@tencent.com}

\author{Qiang Yan}
 \affiliation{
 \institution{WeChat, Tencent}
 \city{Guangzhou}
 \country{China}
}
\email{rolanyan@tencent.com}

\author{Pengjie Ren}
\affiliation{%
  \institution{Shandong University}
  \city{Qingdao}
  \country{China}
}  
\email{p.ren@uva.nl}
\renewcommand{\shortauthors}{Wang et al.}

\acmSubmissionID{756}

\begin{document}
% \title[Few-Shot Electronic Health Record Coding]{Few-Shot Electronic Health Record Coding \\ through Graph Contrastive Learning}
\title{Paying More Attention to Self-attention: Improving Pre-trained Language Models via Attention Guiding}

\begin{abstract}
\Acp{PLM} have demonstrated their effectiveness for a broad range of information retrieval and natural language processing tasks.
As the core part of \acp{PLM}, multi-head self-attention is appealing for its ability to jointly attend to information from different positions.
However, researchers have found that \acp{PLM} always exhibit fixed attention patterns regardless of the input (e.g., excessively paying attention to `[CLS]' or `[SEP]'), which we argue might neglect important information in the other positions.
In this work, we propose a simple yet effective attention guiding mechanism to improve the performance of \acp{PLM} through encouraging the attention towards the established goals.
Specifically, we propose two kinds of attention guiding methods, i.e., the \ac{MDG} and the \ac{PDG}. 
The former definitely encourages the diversity among multiple self-attention heads to jointly attend to information from different representation subspaces, while the latter encourages self-attention to attend to as many different positions of the input as possible.
We conduct experiments with multiple general pre-trained models (i.e., BERT, ALBERT, and Roberta) and domain-specific pre-trained models (i.e., BioBERT, ClinicalBERT, BlueBert, and SciBERT) on three benchmark datasets (i.e., MultiNLI, MedNLI, and Cross-genre-IR).
Extensive experimental results demonstrate that our proposed \ac{MDG} and \ac{PDG} bring stable performance improvements on all datasets with high efficiency and low cost.
\end{abstract}

\maketitle

\acresetall

\section{Introduction}
\Acfp{PLM} have led to tremendous performance increase in a wide range of downstream tasks, including machine translation~\cite{DBLP:conf/wmt/OttEGA18}, text classification~\cite{DBLP:conf/cncl/SunQXH19}, document ranking~\cite{DBLP:journals/corr/abs-1904-07531}, etc.
The core component of \acfp{PLM} is the self-attention mechanism, which allows the model to capture long-range dependency information.

Recently, many studies focus on analyzing the self-attention mechanism, i.e., the weights and connections of attention, to interpret the network or revealing the characteristic of \acp{PLM} \cite{DBLP:conf/blackboxnlp/ClarkKLM19, DBLP:conf/emnlp/KovalevaRRR19, DBLP:conf/blackboxnlp/VigB19}.
These exploration works have found a common phenomenon: despite the success of the self-attention mechanism, these language models exhibit simple attention patterns~\cite{raganato2018analysis,DBLP:conf/acl/VoitaTMST19}.
For example, \citet{DBLP:conf/naacl/DevlinCLT19} and \citet{DBLP:conf/emnlp/KovalevaRRR19} report the phenomenon that 40\% of heads in a pre-trained BERT model simply pay attention to the delimiters, such as `[CLS]' and/or `[SEP].'
Moreover, \citet{DBLP:conf/nips/MichelLN19} demonstrate that multi-headed attentions in WMT \cite{DBLP:conf/nips/VaswaniSPUJGKP17} and BERT \cite{DBLP:conf/naacl/DevlinCLT19} are not necessary to obtain competitive performance.
Likewise, \citet{DBLP:conf/emnlp/RaganatoST20} also confirm that most attentive connections in the encoder do not need to be learned at all, because most self-attention patterns learned by the transformer architecture merely reflect the positional encoding of contextual information.

The effectiveness of self-attention can be improved by introducing a variety of information.
For example, \citet{DBLP:conf/emnlp/LiTYLZ18} demonstrate that the downstream task can be improved by increasing the diversity of attention heads.
Besides, many researches focus on modifying self-attention through external information such as syntactic supervision \cite{DBLP:conf/www/XiaWTC21, DBLP:journals/corr/abs-2012-14642, li2020improving} to improve the input representation.
Their results suggest that adding additional information does help \acp{PLM} improve the effectiveness of downstream tasks. 
However, since these methods modify the  computational process of self-attention, they must re-train the \acp{PLM} from scratch. 
As we all know, training a \ac{PLM} with a large amount of data from scratch will take a lot of computing resources and time.
In addition, extracting additional information, such as syntactic structure, will further increase the computational burden.

\begin{figure}[h]
\centering  
\includegraphics[width=0.45\textwidth]{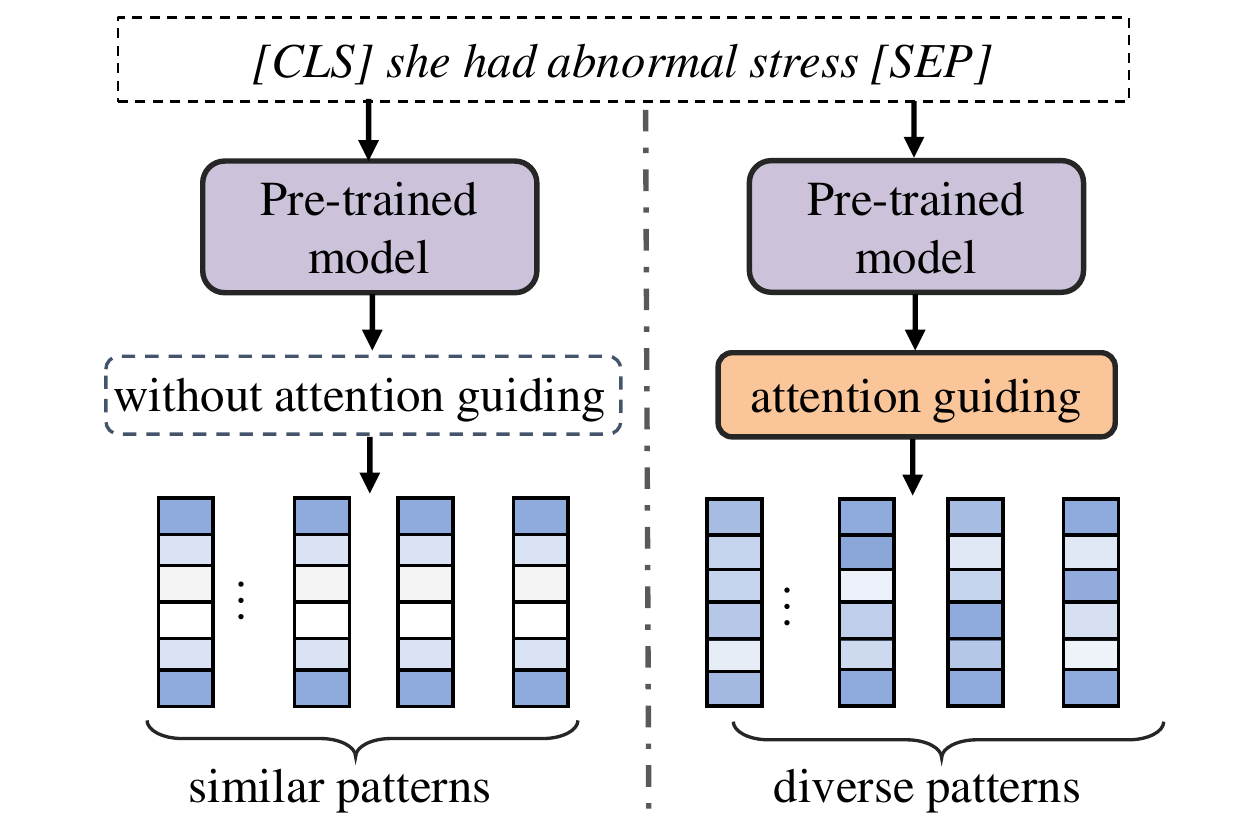}  
\caption{Illustration of attention guiding mechanism.
The learned attention heads always exhibits fixed and similar patterns, which might neglect important information. 
Attention guiding aims to guide the learned attention heads to attend to different parts of the inputs so that more important information can be taken into account.
% \todo{这个图不是为了解释清楚APG和ADG是怎么做的，而是为了通过例子形象解释：1. the learned attention fromPLMs always076focuses on specific delimiters, e.g. ‘[CLS]’; 2. after being guided by APG, the self-attention078begins to attend to positions containing some linguistic information; 3. the guided attention afterADGcan attend to the085broader and diverse positions.}
}
\label{intro}
\end{figure}

Therefore, we seek to investigate the following research question in this paper: is it possible to guide self-attention without extra information in the fine-tuning phrase to improve the performance of downstream tasks?
As shown in Figure~\ref{intro}, the learned attention heads from \acp{PLM} without guiding always present similar patterns, e.g., different heads attend to similar positions.
On the contrast, we seek to design an attention guiding mechanism so that comprehensive and diverse information can be taken into account.
We expect the attention guiding mechanism acts as auxiliary objective to regularize the fine-tuning of downstream tasks.
A similar work is done in \cite{DBLP:conf/emnlp/DeshpandeN20}.
They use several pre-defined attention patterns to guide the training of \acp{PLM}.
Specifically, they add an auxiliary loss to guide the self-attention heads towards a set of pre-defined patterns (i.e., `[Next]', `[Prev]', `[First]', `[Delim]', and `[Period]').
Since these pre-defined patterns only cover a few fixed patterns and cannot introduce more information, the proposed method has limited ability in improving the diversity of attention heads.
% Moreover, \todo{according to \cite{xx},} the self-attention heads need more diversity guidance to dig out more comprehensive and diverse information, rather than frequent attention patterns solely. 
Therefore, in this work, we propose to explore the self-attention guiding methods without pre-defining attention patterns or extra knowledge about the input to encourage the diversity among multiple attention heads.

% For example, \citet{DBLP:conf/naacl/DevlinCLT19} and \citet{DBLP:conf/emnlp/KovalevaRRR19} report the phenomenon that 40\% of heads in a pre-trained BERT model simply pay attention to the delimiters, such as `[CLS]' and/or `[SEP].'
% Based on this observation, \citet{DBLP:conf/emnlp/DeshpandeN20} use several pre-defined attention patterns to guide the training of \acp{PLM}.
% Specifically, they add an auxiliary loss to guide the self-attention heads towards a set of pre-defined patterns (i.e. `[Next]', `[Prev]', `[First]', `[Delim]', and `[Period]'). 
% These patterns are all linguistics-independent.
% However, previous work~\cite{DBLP:conf/blackboxnlp/VigB19,DBLP:conf/blackboxnlp/ClarkKLM19} demonstrate that the substantial syntactic information is also noticed  by \acp{PLM}' self-attention.
% \citet{DBLP:conf/acl/LiZLXC21} and \citet{DBLP:journals/corr/abs-2012-14642} propose to revise the attention score with syntactic structure information, which is shown beneficial for the downstream tasks.
% Moreover, \citet{DBLP:conf/emnlp/LiTYLZ18} demonstrate that the performance on downstream tasks is generally better when the attentions are more diverse.
% Therefore, we believe that self-attention heads need more guidance from linguistic knowledge so that to dig out more comprehensive and diverse information, rather than frequent attention patterns solely.
% % A natural idea arises: can \acp{PLM} be guided towards comprehensive and diverse attention patterns during fine-turning?
% %

We propose an attention guiding mechanism to improve the performance of \acp{PLM} by regularizing its self-attention by explicitly encourage the diversity among multiple attention heads.
Specifically, we propose two kinds of attention guiding methods, i.e., the \ac{MDG} and the \ac{PDG}. 
% % 介绍AMD基本原理即可
The former is used to encourage self-attention to attend to the information from different aspects of the inputs by diverse attention maps.
An attention map is distinctive in its own right, and each could differ significantly from other attention maps \cite{DBLP:conf/iccv/MalisiewiczGE11} so that it can capture the information of input differently with others.
The latter is used to improve the diversity of attention patterns by encouraging self-attention to pay attention to more diverse positions by reducing the correlations with different attention patterns.
We validate the effectiveness of the attention guiding mechanism on multiple general and domain-specific \acp{PLM} by conducting experiments on three benchmark datasets.
Especially, we found that the proposed attention guiding mechanism is still effective on small-scale datasets, demonstrating its significance for low-resource settings.

Our main contributions are as follows:
 \begin{itemize}
     \item We propose two self-attention guiding terms, i.e., \ac{MDG} and \ac{PDG}, for guiding self-attention heads which enable \acp{PLM} to learn comprehensive and diverse attention patterns.
     \item We demonstrate the effectiveness of the attention guiding mechanism on seven general and domain-specific \acp{PLM} across three different datasets and tasks.
 \end{itemize}
\section{Methodology}
\label{sec:method}
%\todo{This section needs more references}

\subsection{Tasks}
In this work, we take the following two different tasks as applications.

\header{Task 1: Natural Language Inference.}
The goal of this task is to predict whether a given hypothesis can be inferred from a given promise.
This task is formulated as a multi-class classification task.
In the implementation, we add a classification layer on top of the `[CLS]' representation derived from the output of the last layer of \acp{PLM}, like most methods~\cite{DBLP:conf/emnlp/ReimersG19,moons2020comparison,DBLP:conf/sigir/YatesNL21}.
The \acp{PLM} are fine-tuned via minimizing the multi-class cross-entropy loss, which is defined as follows:
\begin{equation}
 Loss_{t} = - \frac{1}{|D|} \sum_{i}^{|D|} log(p({y_i}|x_i;\theta)),  
\label{eq:mc-loss}
\end{equation}
where $\theta$  denotes all trainable parameters in the \ac{PLM} and the classification layer, $|D|$ is the number of training samples, and ${y_i}$ is the ground truth for the $i$-th sample $x_i$.

\header{Task 2: Across Medical Genres Querying.}
The objective of this task is to find the research publication that supports the primary claim made in a health-related news article.
This task is formulated as a binary classification task.
Similarly, we add a classification layer on top of the `[CLS]' representation derived from the output of the last layer of \acp{PLM}.
The \acp{PLM} are fine-tuned via minimizing the binary cross-entropy loss as follows:
\begin{equation}
    Loss_{t} = -\frac{1}{|D|}\sum_{i=1}^{|D|}[-y_ilog(\hat{y}_i)-(1-y_i)log(1-\hat{y}_i)], 
\label{eq:bc-loss}
\end{equation}
where ${y_i}$ is the ground truth for the $i$-th sample $x_i$
and $\hat{y_i} = p({y_i}|x_i;\theta)$ is the probability that $i$-th sample belongs to $y_i$.

\subsection{Multi-head self-attention}

%\todo{introduce PLM briefly?}
A \acfi{PLM} is normally a large-scale and powerful
neural network trained with huge amounts of data samples and computing resources~\cite{DBLP:conf/acl/LaiTN20,DBLP:conf/aaai/ChoudhuryD21}.
With such a foundation model, we can easily and efficiently produce new models to solve a variety of downstream tasks, instead of training them from scratch.
\acp{PLM} rely on multi-head self-attention to capture dependencies between tokens~\cite{DBLP:conf/naacl/DevlinCLT19}.
Given a hidden state $\boldsymbol Z$, multi-head self-attention first projects it linearly into queries $\boldsymbol{Q}_{h}$, keys $\boldsymbol{K}_{h}$, and values $\boldsymbol{V}_{h}$ using parameter matrices $ \boldsymbol W_{h}^Q, \boldsymbol W_{h}^K, \boldsymbol W_{h}^V$, respectively. 
The formulation is as follows:

\begin{equation}
    \boldsymbol{Q}_{h}, \boldsymbol{K}_{h}, \boldsymbol{V}_{h} = \boldsymbol Z \boldsymbol{W}_{h}^Q, \boldsymbol{Z} \boldsymbol{W}_{h}^K,\boldsymbol{Z} \boldsymbol{W}_{h}^V.\\
\end{equation}

Then, the self-attention distribution $\boldsymbol A_{h}$ is computed via scaled dot-product of query $\boldsymbol{Q}_{h}$ and key $\boldsymbol{K}_{h}$.
These weights are assigned to the corresponding value vectors $\boldsymbol{V}_{h}$ to obtain output states $\boldsymbol O_{h}$:

\begin{equation}
\boldsymbol O_{h} = \boldsymbol A_{h} \boldsymbol V_{h} ~~with~~ \boldsymbol A_{h} = softmax(\frac{\boldsymbol Q_{h} \boldsymbol K_{h}^\top} {\sqrt{d_k} } ).\\
\end{equation}

Here $\boldsymbol A_{h}$ is the attention distribution produced by the $h$-th attention head. 
$d_k$ is the hidden size.
Finally, the output states $\boldsymbol O_{h}$ of all heads are concatenated to produce the final states $\boldsymbol Z$.

\begin{figure*}[h]
\centering  
\includegraphics[width=1.0\textwidth]{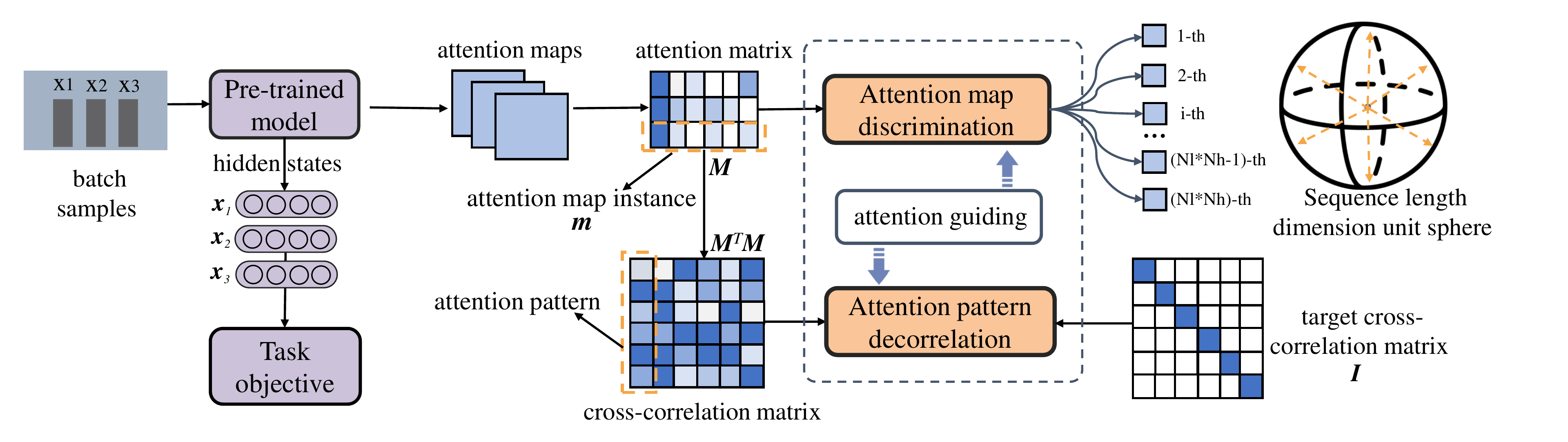}  
\caption{
The pipeline of the proposed \ac{PLM} with attention guiding approach.
The attention guiding mechanism contains \acfi{MDG} and \acfi{PDG}, respectively.
For a batch sample, we use the pre-trained model to encode each sample as a feature vector $ \boldsymbol{x_i}$, and as well as we obtain the attention matrix $ \boldsymbol{M} $ for each sample.  
The optimal feature embedding is learned via task objective plus the losses produced by two attention guiding methods, i.e., \ac{MDG} and \ac{PDG}. 
Both \ac{MDG} and \ac{PDG} try to maximally scatter the attention heads of training samples over the sequence length unit sphere.
%\todo{mark sequence-based prior attention, syntax-based prior attention and synonym-based prior attention in the figure?}
%\todo{no ref. 这个图没有引用}
}
\label{model}
\end{figure*}
%
% To guide the self-attention in the \acp{PLM} towards comprehensive and diverse objectives, we devise a lightweight \ac{AG} method that is pictured in Figure \ref{model}.
% We first describe the framework of \acp{PLM} with \ac{AG} mechanism and then details two attention guiding terms,  i.e. \ac{APG} and \ac{ADG} in \S\ref{sec:ag}. 

\subsection{Attention guiding}
\label{sec:ag}
%\todo{引用图2，先整体介绍一下方法的思路，overall pipeline}
Figure~\ref{model} shows the workflow of our methodology.
Firstly, we use the pre-trained model to encode each sample $x_i$ and obtain its corresponding attention matrix $\boldsymbol{M}$. 
Then the task-objective loss is calculated based on sample representation $\boldsymbol{x}_i$. 
Meanwhile, the attention matrix $\boldsymbol{M}$ will be guided by two terms, i.e., \acf{MDG} and \acf{PDG}, which aim to push the diversity of self-attention and further improve the performance of the downstream task. 
More specifically, for both of downstream tasks, besides the objectives, we also introduce two auxiliary objectives to guide the self-attention.
Formally, the training objective of each task is revised as:
\begin{equation}
\label{losses}
   Loss_{total} = Loss_{t} + \alpha Loss_{mdg} + \beta Loss_{pdg},
\end{equation}
where $L_{total}$ is the ultimate training goal, and it contains three parts of training loss.
$Loss_{t}$ represents the task object, which varies with the task.
$Loss_{mdg}$ denotes the \acf{MDG} term, and $loss_{pdg}$ denotes the \acf{PDG} term. 
These two terms can be either used individually or in combination and they are regulated using $\alpha$ and $\beta$ respectively.
Note that the introduced regularization terms work like $L_1$ and $L_2$ terms which don't introduce any new parameters and only influence the fine-tuning of the standard model parameters.

\subsubsection{\bf{Attention Map Discrimination}}

For a given sample $x_i$, firstly, we need to get the representation of this sample and its corresponding attention maps from the output of the pre-trained models, such as BERT.
The formula can be described as:
\begin{equation}
    \boldsymbol{h}_i, \{\boldsymbol{A}_1, \boldsymbol{A}_2, \boldsymbol{A}_i,...,\boldsymbol{A}_{N_l*N_h}\} = PLM (xi|\theta),
\end{equation}
where $\boldsymbol{h}_i$ denotes the hidden state of `[CLS]' token and we regard it as the representation of sample $x_i$.
 $\{\boldsymbol{A}_1, \boldsymbol{A}_2, \boldsymbol{A}_i,..., \boldsymbol{A}_{N_l*N_h}\}$ is the set of multi-layer multi-head attention maps. 
$\boldsymbol{A}_i$ is the $i$-th attention map, and there are   $N_l \ast N_h$ attention maps produced by the pre-trained model.
$N_l$ and $N_h$ denote the layer number, and head number in each layer, respectively.

Then,  we add a classification layer on top of the `[CLS]' representation $\boldsymbol{h}_i$.
Formally, the formula of the classification layer is implemented by:
\begin{equation}
    \boldsymbol{x}_i = \boldsymbol{W} Relu(\boldsymbol{h}_i),
\end{equation}
where $\boldsymbol{W}$ is the weight matrix, $Relu(\cdot )$ is the activation function, and $\boldsymbol{x}_i$ is the final representation of the given sample. 
By Eq.\ref{eq:mc-loss} or Eq.\ref{eq:bc-loss} which is determined by the downstream task, we can calculate the task-objective loss $L_t$ with the supervision from the ground-truth label of sample $x_i$.

%We pick all layers of \acp{PLM}, obtain their self-attention maps $\{\boldsymbol{A}_1, \boldsymbol{A}_2, \boldsymbol{A}_i,...,\boldsymbol{A}_{N_l*N_h}\}$, and $i$-th attention map is denoted as $\boldsymbol A_i \in \mathbb{R}^{L \times L} $, where $L$ denotes the sequence length.
To simplify the calculation, each attention map $\boldsymbol A_i \in \mathbb{R}^{L \times L} $ is processed as one vector $\boldsymbol m_i \in \mathbb{R}^{L} $ by summing up the attention values that all tokens received.
The corresponding formula of transforming the attention map $\boldsymbol{A}_i$ to the attention vector $\boldsymbol{m}_i$ is: 
\begin{equation}
    \boldsymbol{m}_i = \sum_{j}\boldsymbol{A}_{i,j},
\end{equation}
where $i$ represents the $i$-th attention map and $j$ is the column index of the attention map $\boldsymbol{A}_i$.

Since the self-attention mechanism in \ac{PLM} is multi-layer multi-head architecture, there are multiple attention vectors are produced.
we organize all the attention vectors,  into a matrix $\boldsymbol M \in \mathbb{R}^{ (N_l \ast  N_h) \times L}$.
Specifically, we concatenate all the attention vector $m_i$ to construct the attention matrix $\boldsymbol{M}$.
Formally, the corresponding formula is as follows:
\begin{equation}
 \boldsymbol{M} =  \boldsymbol{m}_1 \oplus  \boldsymbol{m}_2 \oplus \boldsymbol{m}_i,..., \oplus \boldsymbol{m}_{N_l \ast N_h},
\end{equation}
where  $\oplus $ denotes the concatenate operation and $\boldsymbol M \in \mathbb{R}^{ (N_l \ast  N_h) \times L} $ represents the attention matrix.

Inspired by \cite{DBLP:conf/cvpr/WuXYL18, DBLP:conf/iclr/TaoTN21}, we apply the instance discrimination method to push the diversity of attention maps so that the rich information of the input can be captured. 
The objective function is formulated based on the softmax criterion. 
Each attention map is assumed to represent a distinct class.
That is, attention map $\boldsymbol{m}_i \in \mathbb{R}^L $, i.e. the $i$-row of the attention matrix $\boldsymbol{M}$, is classified into the $i$-th class. 
Accordingly, the weight vector for the $i$-th class can be approximated by a vector $\boldsymbol{m}_i$.
The probability of one attention map $\boldsymbol{m}$ being assigned into the $i$-th class is: 
\begin{equation}
    P(i|\boldsymbol{m}) = \frac{exp(\boldsymbol{m}_i^\top \boldsymbol{m}/\tau   )}{ {\textstyle \sum_{j=1}^{N_l \ast N_h} exp(\boldsymbol{m}_j^\top \boldsymbol{m}/\tau )} }, 
\end{equation}
where $\boldsymbol{m}_j^\top \boldsymbol{m}$ measures how well $\boldsymbol{m}$ matches the $j$-th class because $\boldsymbol{m}_j$ is regarded as the weight of $j$-th class. 
$\tau$ is a temperature parameter that controls the concentration of the distribution \cite{DBLP:journals/corr/HintonVD15}, and $m$ is normalized to $\left \| m \right \|=1$.
The objective maximizes the joint probability $ {\textstyle \prod_{i=1}^{N_l\ast N_h}}P_\theta(i|f_\theta (\boldsymbol{m}_i)) $ as 
\begin{equation}
\begin{split}
    Loss_{mdg} &= -\sum_{i=1}^{N_l \ast N_h} logP(i|f_\theta (\boldsymbol{m}_i)),\\
    &= - \sum_{i}^{N_l \ast N_h} log(\frac{exp(\boldsymbol{m}_i^ \top \boldsymbol{m}_i/\tau)}{{ \sum_{j=1}^{N_l \ast N_h}exp(\boldsymbol{m}_j ^ \top \boldsymbol{m}_i/\tau)} } ).
\end{split}
\label{loss_mdg}
\end{equation}

\subsubsection{\bf{Attention Pattern Decorrelation}}
We have analyzed that the multi-head attention heads are likely to suffer from the redundancy problem where each attention vector focuses on a very similar region.
To encourage each attention head to capture the information from different positions of the inputs,  at the same time, we propose another attention guiding term, i.e., the \acfi{PDG}.

Inspired by ~\cite{DBLP:conf/aaai/Li0LPZ021}, we regard the $l$-th column of $\boldsymbol M$, i.e., $\boldsymbol M^\top_l  \in \mathbb{R}^{N_l \ast N_h} $ as the soft representation of the $l$-th attention pattern.  
Conventionally, attention patterns should be independent to ensure that redundant information is reduced.
The objective function is applied to push the diversity of attention patterns and reduce pattern redundancy,  which tries to make the cross-correlation matrix computed from the attention matrix as close to the identity matrix as possible. 

The formula of \ac{PDG} term aims to construct independent attention patterns and is as follows:
\begin{equation}
    Loss_{pdg} = \left \| \boldsymbol M^\top \boldsymbol M - \boldsymbol I \right \|_{F} ^2,
\label{loss_pdg}
\end{equation}
where $\boldsymbol M \in \mathbb{R}^{(N_l \ast N_h) \times L}$ is the attention matrix, $\boldsymbol I \in \mathbb{R}^{L \times L}$ is the identity matrix and $||.||^2_F$ denotes the squared Frobenius Norm~\cite{DBLP:journals/ijcm/YuanYLJ20}.  %~\cite{DBLP:journals/ijcm/YuanYLJ20}
$\boldsymbol M^\top \boldsymbol M \in \mathbb{R}^{L \times L} $ can be seen as the cross-correlation matrix of different attention patterns. 
Minimizing the difference between the cross-correlation matrix and the identity matrix $\boldsymbol I$ is equivalent to making the attention patterns diverse so that they focus on different tokens~\cite{DBLP:conf/wacv/ParkHKY20,DBLP:conf/icml/ZbontarJMLD21}.

% The effect of \ac{ADG} term will be analyzed in section \ref{sec:analysis}.

\section{Experimental Setup}
\label{sec:experiment}
To evaluate the effectiveness of our proposed attention guiding mechanism, we conduct extensive experiments of a variety of pre-trained models on different downstream tasks.
We demonstrate that the attention guiding mechanism can promote the diversity of attention heads and further improve the performance on different downstream tasks.

\subsection{Datasets and evaluation}
\begin{table*}[]
\centering
\caption{Performance comparison (\%) of \acp{PLM} with or without attention guiding.
\textbf{Bold face} indicates the improved performance with \ac{AG} in terms of the corresponding metrics.
$^\ast$~~ means $p < 0.05$ and $^{\ast \ast}$~~means $p < 0.01$ in t-test, respectively.
SOTA represents the best performing methods by Nov. 2021, to the best of our knowledge.
The matched test set of MultiNLI for the experiments. 
}
%Significant improvements over the best baseline results are marked with $^\ast$ (t-test, $p < 0.05$).}
\begin{threeparttable}
\setlength\tabcolsep{3pt}%调列距
\begin{tabular}{l | cccc| cccc| ccccc}
\toprule
\multirow{2}{*}{\bf Methods}&\multicolumn{4}{c}{\bf MultiNLI} &\multicolumn{4}{c}{\bf MedNLI} &  \multicolumn{5}{c}{\bf Cross-genre-IR}
\\
 \cmidrule(lr){2-5} \cmidrule(lr){6-9} \cmidrule(lr){10-14} 
 & ACC &Precision &Recall &F1 & ACC &Precision &Recall &F1 & MRR  & R@1 & R@3 & R@5 & R@20
\\
\midrule
SOTA & 87.90 & -- & -- & -- & 84.00 & -- & -- & -- & 69.50 & 62.50 & 74.30 & 78.30 & 82.80 \\
\midrule

BERT & 83.24 &83.19 &83.17 & 83.17 &76.02 &76.25 &76.01 &76.11& 73.18 & 62.84 & 80.38 & 85.07 & 94.58\\ 
BERT+AG &  \bf{83.89} & \bf{83.91} & \bf{83.82} & \bf{83.83} & \bf{77.22}$^\ast$ &\bf{77.35}$^\ast$ & \bf{77.24}$^\ast$ &\bf{77.26}$^\ast$ & \bf{73.38} & \bf{63.69} & \bf{80.74} & 84.59 & 94.09\\ 			
\midrule
ALBERT &83.09 &83.05 & 83.04 &83.02 & 77.76 &78.04 &77.78 &77.84 & 71.00 & 62.06 & 76.07 & 80.02& 93.21
 \\
ALBERT+AG & \bf{84.41}$^{\ast \ast}$ & \bf{84.44}$^{\ast \ast}$ & \bf{84.37}$^{\ast \ast}$ & \bf{84.37}$^{\ast \ast}$ & \bf{78.76}$^{\ast \ast}$ & \bf{78.72}$^{\ast\ast}$ & \bf{78.80}$^{\ast\ast}$ & \bf {78.74}$^{\ast\ast}$ & \bf{71.86}$^\ast$ & 62.06 &  \bf{78.08}$^\ast$ & \bf{82.95}$^\ast$ & \bf{93.42}$^\ast$ \\
\midrule
Roberta & 85.95 &85.91 &85.94 &85.91 & 80.17 &80.12 &80.12 &80.14 & 78.67 & 71.04 & 83.58 & 86.85 & 94.86\\
Roberta+AG & \bf{86.77}$^{\ast}$ & \bf{86.67}$^\ast$ &\bf{86.72}$^\ast$ & \bf{86.67}$^\ast$ & \bf{80.31} & \bf{80.36} & \bf{80.31} & \bf{80.32} &  \bf{79.88}$^\ast$ & \bf{72.05}$^\ast$ & \bf{84.47}$^\ast$ & \bf{87.70}$^\ast$ & \bf{95.68}$^\ast$ \\		
\midrule

BioBERT& 81.37 &81.35 &81.33 & 81.28 &81.86 &81.81 &81.84 &81.83 & 83.72 &76.48 & 88.83 & 92.52 & 97.21\\
BioBERT+AG &\bf{81.54}$^{\ast \ast}$ & \bf{81.46}$^{\ast \ast}$ & \bf{81.47}$^{\ast \ast}$ & \bf{81.45}$^{\ast \ast}$ & \bf{82.63} & \bf{82.68} & \bf{82.65} & \bf{82.64} & \bf{83.76} &75.87 &\bf{90.17} & \bf{92.88} &96.84\\
\midrule 
ClinicalBERT & 79.68 & 79.81 & 79.69 &79.66 &80.73& 80.67& 80.71 &80.66 & 71.83 & 61.99 & 78.30 & 83.58 & 92.03\\  
ClinicalBERT+AG & \bf{80.07}$^{\ast \ast}$ & \bf{80.13}$^{\ast \ast}$ & \bf{80.07}$^{\ast \ast}$ & \bf{80.09}$^{\ast \ast}$ & \bf{82.63}$^\ast$ & \bf{82.68}$^\ast$ & \bf{82.67}$^\ast$ & \bf{82.64}$^\ast$ & \bf{72.02}$^\ast$ & \bf{62.24}$^\ast$ & 78.18 & \bf{83.70}$^\ast$ & 91.91\\			
\midrule 
BlueBERT   & 79.13  & 79.10 & 79.10 & 79.06 &83.90 &83.94 &83.91 &83.92 & 69.68 & 59.65 & 75.67 & 82.08 & 91.75\\
BlueBERT+AG & \bf{79.21}$^\ast$ & \bf{79.17}$^\ast$ &\bf{79.17}$^\ast$ & \bf{79.13}$^\ast$ & \bf{84.32}$^{\ast\ast}$ & \bf{84.46}$^{\ast\ast}$ & \bf{84.33}$^{\ast\ast}$ & \bf{84.34}$^{\ast\ast}$ & \bf{71.02}$^\ast$ & \bf{61.47}$^\ast$ & \bf{77.61}$^\ast$ & 81.53 & 91.38\\ 	
\midrule
SciBERT & 83.61 &83.68 & 83.56 & 83.58 &79.47 &79.42 &79.47 &79.43 & 80.72 & 72.80 & 86.65 & 90.57 & 96.48\\ 
SciBERT+AG & \bf{83.74}$^\ast$ & \bf{83.79}$^\ast$ & \bf{83.70}$^\ast$ & \bf{83.71}$^\ast$ & \bf{80.03}$^\ast$ & \bf{80.03}$^\ast$ &\bf{80.11}$^\ast$ & \bf{83.01}$^\ast$ & \bf{81.67}$^\ast$ & \bf{72.84}$^\ast$ & \bf{88.07}$^\ast$ & \bf{91.50}$^\ast$ & \bf{97.41}$^\ast$\\			
\bottomrule       
\end{tabular}
\end{threeparttable} 
\label{result} 
\end{table*}

We conduct experiments on the following datasets.

\begin{itemize}[leftmargin=*]
\item \textbf{MultiNLI}\footnote{\url{https://cims.nyu.edu/~sbowman/multinli/}}~\cite{DBLP:conf/naacl/WilliamsNB18} is a crowd-sourced collection of 433k sentence pairs annotated with textual entailment information, i.e., entailment, contradiction, and neutral.
This dataset is for the natural language inference task, which is also popular for evaluating various \acp{PLM}~\cite{DBLP:conf/emnlp/DeshpandeN20,DBLP:conf/acl/WangBHDW21}.
Accuracy (ACC for short) is standard metric on this task.
At the same time, we also report other metrics commonly used in the classification tasks, such as Precision, Recall and F1.
\item \textbf{MedNLI}\footnote{\url{https://physionet.org/content/mednli/1.0.0/}} ~\cite{DBLP:conf/emnlp/RomanovS18} is  for natural language inference in clinical domain, which has the same data structure as MultiNLI.
Accuracy is also the standard metric on this dataset.
Like on the MultiNLI, we also report the Precision, Recall and F1.
\item \textbf{Cross-genre-IR}\footnote{\url{https://github.com/chzuo/emnlp2020-cross-genre-IR}} ~\cite{DBLP:conf/emnlp/ZuoAB20} is for the across medical genres querying task, where each claim (i.e., he news headline) is associated with at least one peer-reviewed research publication supporting it. 
For each claim, it needs to re-rank the candidate publications to obtain the correct ones.
Following the original authors, we report the Mean Reciprocal Rank (i.e., MRR) and Recall@K (i.e., R@K = 1, 3, 5, 20) metrics.
\end{itemize}

\subsection{\acp{PLM} for comparison}
We consider seven transformer-based \acp{PLM}: three are pre-trained over general language corpora (BERT, ALBERT, and Roberta) and four are pre-trained over biomedical corpora (BioBERT, ClinicalBERT, BlueBert, and SciBERT).

\begin{itemize}[leftmargin=*]
\item \textbf{BERT}\footnote{\url{https://huggingface.co/bert-base-uncased}}~\cite{DBLP:conf/naacl/DevlinCLT19} is a multi-layer bidirectional Transformer encoder.
Since the following versions of the \acp{PLM} are often based on the BERT-base-uncased version (12 layers and 768 hidden size with 108M parameters), we use the BERT-base-uncased here for a fair comparison. 

\item \textbf{Roberta}\footnote{\url{https://huggingface.co/roberta-base}}~\cite{DBLP:journals/corr/abs-1907-11692} has the same architecture as BERT, but with a lot of changes on the training mechanism, such as a more random mask mechanism.
We use the Roberta-base here for comparison. 

\item \textbf{ALBERT}\footnote{\url{https://huggingface.co/albert-base-v2}}~\cite{DBLP:conf/iclr/LanCGGSS20} compresses the architecture of BERT by factorized embedding parameterization and cross-layer parameter sharing.
We use the ALBERT-base-v2 version.

\item \textbf{BioBERT}\footnote{\url{https://huggingface.co/dmis-lab/biobert-v1.1}}~\cite{DBLP:journals/bioinformatics/LeeYKKKSK20} is the first BERT pre-trained on biomedical corpora.
It is initialized with BERT's pre-trained parameters and then further pre-trained over PubMed abstracts and full-text articles.
We use the best version BioBERT V1.1.

\item \textbf{ClinicalBERT}\footnote{\url{https://huggingface.co/emilyalsentzer/Bio_ClinicalBERT}}~\cite{DBLP:journals/corr/abs-1904-03323} is initialized from BioBert v1.0 and further pre-trained over approximately 2 million notes in the MIMIC-III v1.4 database. 

\item \textbf{BlueBERT}\footnote{\url{https://huggingface.co/ttumyche/bluebert}}~\cite{DBLP:conf/bionlp/PengYL19} is firstly initialized from BERT  and further pre-trained over biomedical corpus of PubMed and clinical notes.

\item \textbf{SciBERT}\footnote{\url{https://huggingface.co/nfliu/scibert_basevocab_uncased}}~\cite{DBLP:conf/emnlp/BeltagyLC19} is a BERT-base model pre-trained on 1.4M papers from the semantic scholar, with 18\% of papers from the computer science and 82\% from the biomedical domain.

\item \textbf{SOTA.} We also compare with state-of-the-art methods on each dataset, which are based on Roberta, BlueBERT and BERT, to the best of our knowledge \cite{DBLP:journals/corr/abs-2111-02188, DBLP:conf/bionlp/PengYL19, DBLP:conf/emnlp/ZuoAB20}.
\end{itemize}

\subsection{Implementation details}
The proposed attention guiding mechanism acts on all attention heads out from \acp{PLM}.
We fine-tune all \ac{PLM} models for 5 epochs, 20 epochs, and 5 epochs on the MultiNLI, MedNLI, and Cross-genre-IR datasets, respectively.
The hidden size $d_k$ is 768 and sequence length $L$ is set to 256 of each \ac{PLM}. 
We use the Adam optimizer (learning rate 1e-05) for all models and the batch size is set as the maximum according to the memory of a GeForce RTX 3090 GPU.
Specifically, the batch size of ALBERT is set to 56 and other \acp{PLM} is 64 on different datasets. 
The $\alpha$ and $\beta$ in Eq.~\ref{losses} are selected from the set \{0.1, 0.01, 0.001, 0.0001\} according to grid search.
The temperature parameter in Eq.~\ref{loss_mdg} is set to 1.0.

\subsection{Results on different pre-trained models}
The results of all \acp{PLM} on different tasks are listed in Table \ref{result}.
From the results, we have several observations.

First, the proposed \ac{AG} can improve the performance of the \acp{PLM} on all tasks generally.
For example, on the MultiNLI and MedNLI datasets, all the selected pre-trained models with attention guiding can promote the performance of downstream tasks on all \acp{PLM} in terms of all metrics.
Similarly, on the Cross-genre-IR dataset, most of the metrics of the task can be promoted by our attention guiding method. 
Moreover, the encouraging findings are that simply adding \ac{AG} to BlueBERT (i.e., BlueBERT+\ac{AG}) outperforms SOTA on the MedNLI dataset, and BioBERT+\ac{AG} is far better than SOTA on the Cross-genre-IR dataset.  These figures show the effectiveness of the proposed attention guiding. 
The reason why \ac{AG} works is that the self-attention after guiding has a better ability to attend to broader and more diverse tokens, which benefit for downstream tasks.

Second, \ac{AG} plays different roles on different models and datasets.
For example, the proposed attention mechanism always improves the performances on MultiNLI and MedNLI datasets, while on the Cross-genre-IR dataset, some metrics drop slightly, e.g., R@20 drops from 94.58\% to 94.09\% of BERT after attention guiding. 
Moreover, we also observe that the performances of different \acp{PLM} are always improved in terms of MRR, and more \acp{PLM}  can be improved when K is small in terms of  R@K metrics.
This suggests that the attention guiding method may be influenced by some charismatics of the evaluation metrics.
For instance, the R@20 is difficult to be promoted by attention guiding.
However, although there is a slight decrease in some metrics, in most cases, our attention guiding method can improve the pre-trained model effectively. 

Third, according to our results, \ac{AG} plays a bigger role on small datasets.
For example, the biggest improvements of \ac{AG} reach 3.58\% in terms of F1 and 2.95\% in terms of R@5 on the MedNLI (11k sentence pairs) and  Cross-genre-IR (48k sentence pairs) datasets, respectively, which are greater than these on the MultiNLI (443k sentence pairs) dataset, i.e., 0.98\%.
To further explore this phenomenon, we vary the training size of MedNLI dataset to evaluate \ac{AG}'s role, and the details refer to \S\ref{sec:train_size}.

% We also list state-of-the-art methods on each dataset, which are based on Roberat, BlueBERT, and BERT, to the best of our knowledge, and they are reported by \citet{DBLP:journals/corr/abs-2111-02188, DBLP:conf/bionlp/PengYL19, DBLP:conf/emnlp/ZuoAB20} respectively.

\subsection{Comparison with different attention guiding methods}

\begin{table*}[]
\centering
\caption{Performance comparison (\%) of \acp{PLM} with different guiding methods.
Significant improvements over the best baseline results are marked with $^\ast$ (t-test, $p < 0.05$).}
\begin{threeparttable}
% \resizebox{0.9\textwidth}{!}{
\begin{tabular}{l | cccc| cccc| ccccc}
\toprule
\multirow{2}{*}{\bf Methods}&\multicolumn{4}{c}{\bf MultiNLI} &\multicolumn{4}{c}{\bf MedNLI} &  \multicolumn{5}{c}{\bf Cross-genre-IR}
\\
 \cmidrule(lr){2-5} \cmidrule(lr){6-9} \cmidrule(lr){10-14} 
& ACC &Precision &Recall &F1 & ACC &Precision &Recall &F1 & MRR  & R@1 & R@3 & R@5 & R@20
\\
\midrule
BERT  & 83.24 & 83.19 & 83.17 & 83.17 & 76.02 & 76.25 & 76.01 & 76.11 & 73.18 & 62.84 & 80.38 & 85.07 & 94.58 \\ 
BERT+AG  & \bf{83.89}$^\ast$ & \bf{83.91}$^\ast$ & \bf{83.82}$^\ast$ & \bf{83.83}$^\ast$ & \bf{77.22}$^\ast$ & 77.35 & \bf{77.24}$^\ast$ & \bf{77.26}$^\ast$ & 73.38 & 63.69 & 80.74 & 84.59 & 94.09\\ 			
\midrule
+[First] & 83.65 & 83.63 & 83.58 & 83.59 & 76.02 & 76.38 & 76.01 & 76.13 & 72.81 & 62.24 & 80.24 & 85.44 & \bf{94.70}\\
 +[Next] & 83.71 & 83.74 & 83.64 & 83.66 & 76.93 & 77.23 & 76.93 & 77.04 & 72.84   & 63.79 & 77.69 & 84.47 & 93.61 \\
 +[Prev] & 83.65 & 83.67 & 8.3.57 & 83.59 & 77.07 & 77.52 & 77.06 & 77.15 & 70.98 &60.60 & 77.53 & 83.39 & 93.37 \\
 +[Period] & 83.70 & 83.71 & 83.64 & 83.65 & 76.02 & 76.38 & 76.01 & 76.13 & 73.85 & 64.40 &  79.87 & 85.56 & 93.73\\
 +[Delim] & 83.65 & 83.67 & 83.59 & 83.60 & 75.67 & 75.71 & 75.66 & 75.69 & 70.97 & 61.69 & 76.31 & 82.16 & 90.86 \\  
\midrule
+[PMI] &83.68 & 83.69 & 83.61 & 83.62 & 77.07 & \bf{77.60} & 77.05 & 77.16 & 72.41  &  62.61    &  79.11 & 84.29    & 93.12
\\
+[Dependency] & 83.59 & 83.61 & 83.53 & 83.55 & 76.37 & 76.61 & 76.37 & 76.46 &73.10 &  64.37 & 78.38 & 82.76  & 92.57
\\
+[WordSim] & 83.73 & 83.74 & 83.67 & 83.68 & 76.72 & 77.03 & 76.72 & 76.81 &\bf{74.15} & \bf{64.48} & \bf{80.91} & \bf{85.80} & 94.22
\\
\bottomrule       
\end{tabular}
\end{threeparttable}  
\label{knowledge} 
\end{table*}
We also study how the attention guiding mechanism compared with other related works. 
Since all the related works need to re-train a new model, rather then our work acts in the fine-tuning phase of the pre-trained model.
Therefore, it is difficult to compare them directly.
For comparison, we implement different methods to guide self-attention during the fine-tuning phase of pre-trained models. 
Specifically, we implemented five kinds of attention guiding patterns (i.e., `[Next]', `[Prev]', `[First]', `[Delim]', and `[Period]') proposed by \citet{DBLP:conf/emnlp/DeshpandeN20}.
In addition, we also implemented three kinds of attention guiding methods based on the knowledge extracted from the input.
 Specifically, the guiding methods based on knowledge are guiding attention heads with co-occurrence relationships, syntactic dependencies, and similar relationships between words/tokens, namely `[PMI]', `[Dependency]' and `[WordSim]'  respectively. 
These prior pieces of knowledge about self-attention we used are proposed by \citet{DBLP:journals/corr/abs-2012-14642} and \citet{DBLP:conf/www/XiaWTC21}.
The results of these different guiding methods are reported in Table \ref{knowledge}, and Table \ref{knowledge} shows that:

First, no matter which kind of guidance method can improve the effectiveness of the pre-trained model, i.e., BERT, in varying degrees. For example, the ACC of BERT on the MultiNLI dataset increases from 83.24\% to 83.73\% after the guidance of word similarity knowledge (i.e., `[WordSim]').
The results also show that self-attention heads need to be guided, as long as the guiding method is reasonable, such as using some fixed common attention patterns or using the knowledge derived from the input, the performance of mainstream tasks can also be improved.

Second, although three types of attention guiding methods, proposed by ours, \citet{DBLP:conf/emnlp/DeshpandeN20}, \citet{DBLP:journals/corr/abs-2012-14642} and \citet{DBLP:conf/www/XiaWTC21} can improve the performances of \acp{PLM} in mainstream tasks, different guiding methods play different roles in different datasets.
Our \ac{AG} is superior to other methods on almost all metrics, such as the ACC on MedNLI and MutiNLI datasets.
But on the Cross-genre-IR datasets, `[WordSim]' method is better than ours in terms of most metrics.
This suggests that the effect of different attention guiding methods may be affected by the dataset, and trying one or more attention guiding methods on a specific dataset may maximize the effectiveness of the pre-trained models.  
How to choose appropriate prior knowledge or guidance methods for self-attention may be a problem that needs further exploration.

\section{Analysis}
\label{sec:analysis}

\begin{table*}[]
\centering
\caption{Performance comparison (\%) of different methods. \textbf{Bold face} indicates the best results in terms of the corresponding metrics.
The matched test set of MultiNLI for the experiments.
Significant improvements over the best baseline results are marked with $^\ast$ (t-test, $p < 0.05$).}
\label{tb:as} 
% \resizebox{0.80\textwidth}{!}{
\begin{tabular}{l  cccc| cccc| ccccc}
\toprule
\multirow{3}{*}{\bf Variants} &\multicolumn{4}{c}{\bf MultiNLI} &\multicolumn{4}{c}{\bf MedNLI} &  \multicolumn{5}{c}{\bf Cross-genre-IR}\\
\cmidrule(lr){2-14}
&\multicolumn{4}{c}{Roberta} &\multicolumn{4}{c}{ BlueBERT} &  \multicolumn{5}{c}{BioBERT}
\\
 \cmidrule(lr){2-5} \cmidrule(lr){6-9} \cmidrule(lr){10-14} 
 & ACC &Precision &Recall & F1  & ACC &Precision & Recall &F1 & MRR & R@1 & R@3 & R@5 & R@20 
\\
\midrule 
Default & \bf{86.77}$^\ast$ & \bf{86.67}$^\ast$ & \bf{86.72}$^\ast$ & \bf{86.67}$^\ast$ & \bf{84.32}$^\ast$ &\bf{84.46}$^\ast$ & \bf{84.33}$^\ast$ & \bf{84.34}$^\ast$ & \bf{83.76} & 75.87 & \bf{90.17} & \bf{92.88} & \bf{96.84}\\		
\midrule
w/o-AG & 85.95 & 85.91 & 85.94 & 85.91 & 83.90 & 83.94 & 83.91 & 83.92 & 83.72 & \bf{76.48} & 88.83 & 92.52 & 97.21\\
w/o-MDG & 86.52 & 86.44 & 86.49 & 86.43 & 84.18 & 84.21 & 84.18 & 84.15 & 83.48     & 75.99  & 88.63 &92.52 &  96.84 \\			
w/o-PDG & 86.57 & 86.48 & 86.52 & 86.48 & 82.98 & 83.15 & 82.98 & 83.02 & 83.28 & 75.63 & 88.83 &   92.76 & 96.72\\
\bottomrule       
\end{tabular}
% }
\end{table*}
\begin{figure*}[ht]
\centering  
\includegraphics[width=0.88\textwidth]{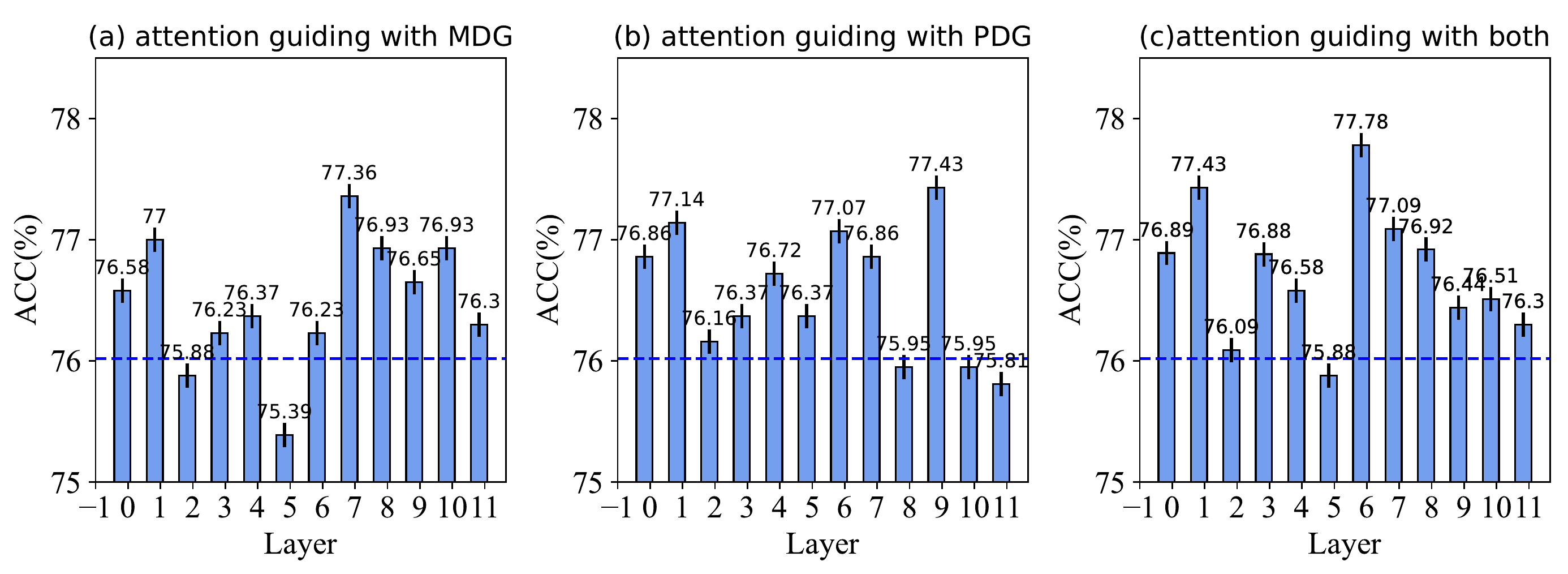}  
\caption{Performances of each BERT layer with the proposed \ac{AG} mechanism.
Figure (a), (b) and (c) demonstrate the results of using \ac{MDG}, \ac{PDG}, and both of \ac{MDG} and \ac{PDG} to guide each layer of BERT, respectively.
}
\label{layer}
\end{figure*}

\begin{figure}[ht]
\centering  
\includegraphics[width=0.45\textwidth]{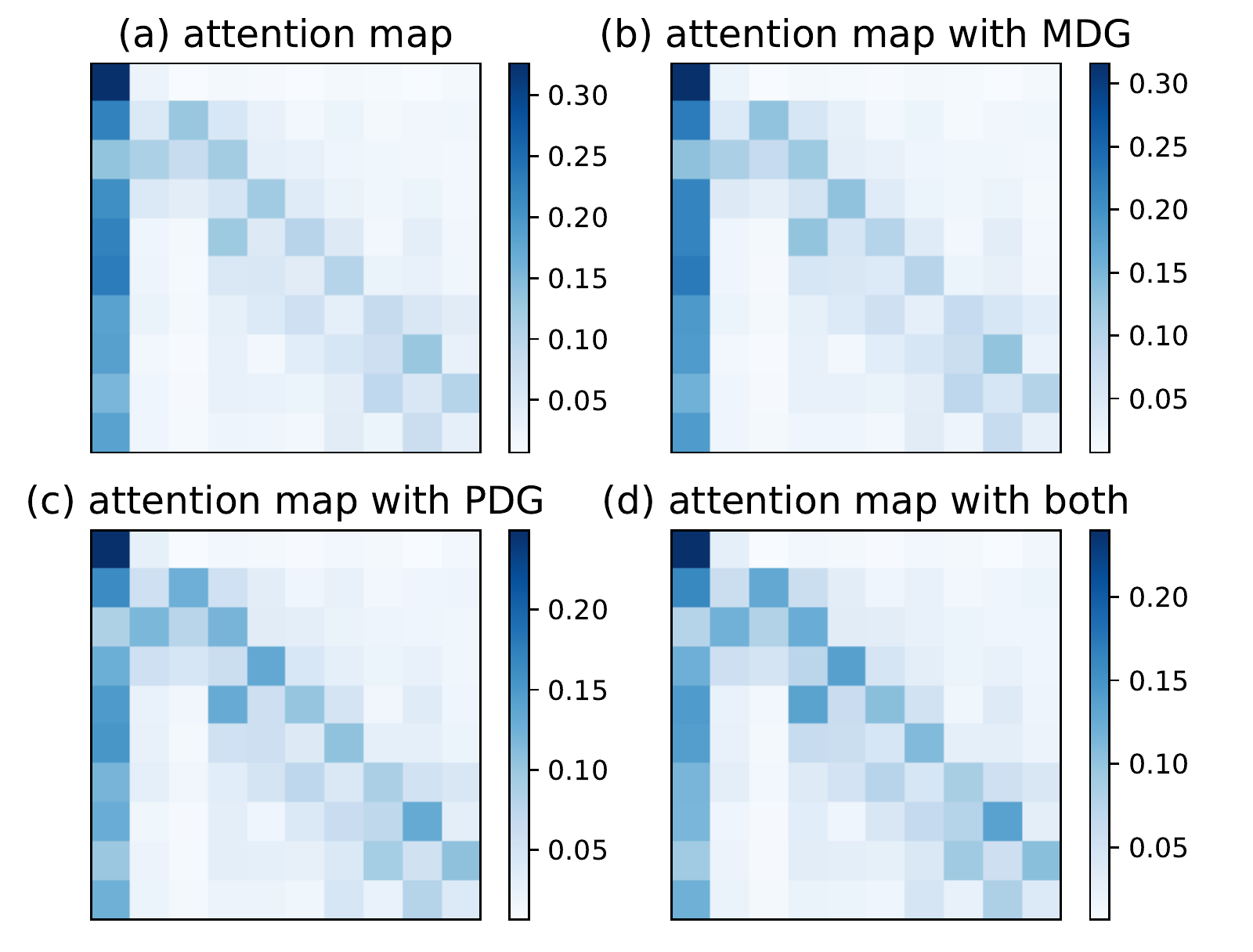}  
\caption{Attention heatmap for a random sample.      
Figure (a) represents attention without \ac{AG}, and Figure (b)-(d) represent attention with \ac{MDG}, with \ac{PDG},  and with both of \ac{MDG} and \ac{PDG}, respectively.}
\label{attention}
\end{figure}

\begin{figure*}[ht]
\centering  
\includegraphics[width=1.0\textwidth]{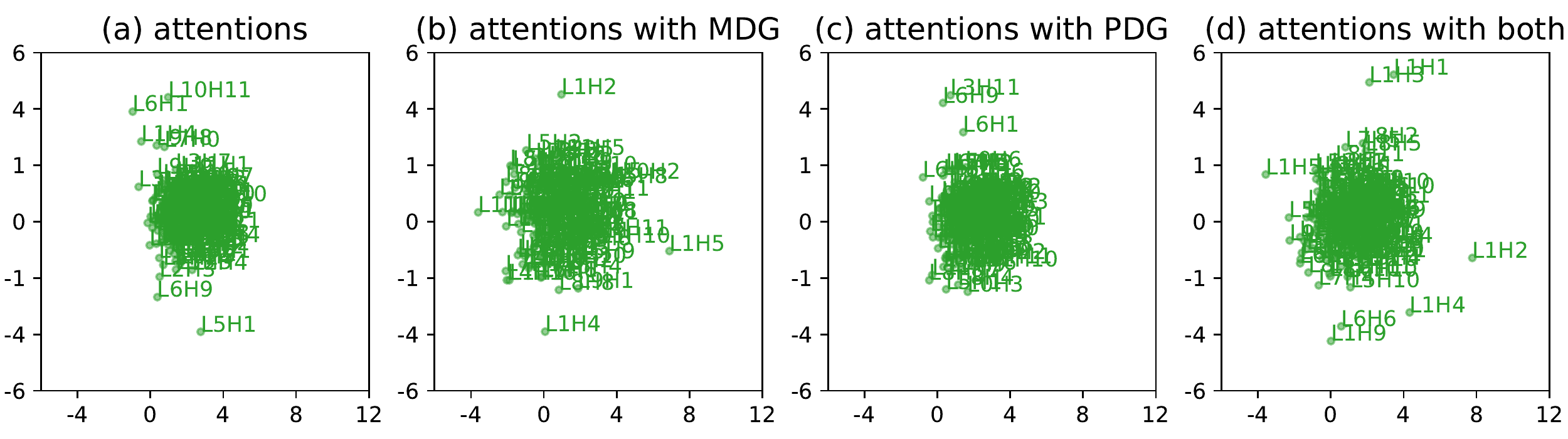}  
\caption{Attention principal component analysis for a random sample.
Figure (a) denotes the spatial distribution of multi-layer multi-head attentions, and Figure (b)-(d) represent the attentions with \ac{MDG}, with \ac{PDG}, and with both of \ac{MDG} and \ac{PDG} respectively.} 
% \todo{为什么用3个sample，一个岂不是更清晰直观？}
% \todo{Prefer Latex subfigure instead of all subfigures in one}}
\label{pca}
\end{figure*}

\subsection{Ablation study}
To analyze where the improvements of \ac{AG} come from, we conduct an ablation study of the best model on each dataset.
Obviously, it is much easier to comprehend that models pre-trained by medical corpus are better at handling medical-related tasks (i.e., MedNLI and Cross-genre-IR).
Therefore, the best pre-trained models on MultiNLI, MedNLI, and Cross-genre-IR datasets are Roberta, BlueBERT, and BioBERT respectively. 
The detailed results are shown in Table \ref{tb:as}.
We consider the following three settings: 
(1) {\bf w/o-AG} denotes \acp{PLM} without \ac{AG}, i.e., baseline \acp{PLM}. 
(2) {\bf w/o-MDG} denotes \acp{PLM} without the \ac{MDG} term but reserving the \ac{PDG} term.
(3) {\bf w/o-PDG} denotes removing the \ac{PDG} term but reserving the \ac{MDG} term.

The results in Table \ref{tb:as} show that \ac{MDG} and \ac{PDG} are helpful for \acp{PLM} as removing either of them leads to a decrease in performance in almost all the metrics. 
Besides, on the MedNLI and Cross-genre-IR datasets, the most obvious declines are the variants removing \ac{PDG}, i.e., w/o-PDG.
This illustrates that the attention pattern decorrelation guiding can bring more valuable attention information for \ac{PLM} because \ac{PDG} can push each head to attend to different positions of the inputs to capture diversity information.

We also note that the R@1 on the cross-genre-IR dataset declines slightly, e.g., the R@1 drops from 76.48\% to 75.87\% with \ac{AG}.
Nevertheless, \ac{AG} is still effective as the other metrics (i.e., MRR, R@3, R@5, and R@20) still get improvements after the \ac{AG} mechanism.

\subsection{Effect of \ac{AG} on different layers}

As the proposed \ac{AG} (i.e., \ac{MDG} and \ac{PDG}) can be applied to any layers of \acp{PLM}, we design experiments to see their effect on different layers of BERT.
The results of BERT with \ac{MDG}, BERT with \ac{PDG}, and BERT with both of \ac{MDG} and \ac{PDG} on different layers are summarized in Figure~\ref{layer}.
The blue dashed line indicates BERT without \ac{AG}.

From Figure~\ref{layer}, we can see that most layers can benefit from \ac{MDG} and \ac{PDG} obviously, such as the ACC increases from 76.02\% to 77.36\% at layer 7 after being guided by \ac{MDG}. 
And similarly, the ACC is improved from 76.02\% to 77.43\% at layer 9 after being guided by \ac{PDG}. 
Moreover, lower and middle layers can always benefit from \ac{PDG}, while for top layers, there are some declines occasionally. 
For example, at layer 11, the ACC drops from 76.02\% to 75.81\% after being guided by \ac{PDG}.
On the contrary, some declines happen at the lower and middle layers of BERT with \ac{MDG}, e.g.,  the ACC of BERT is down 0.63\% at layer 5.
That is understandable as the functions of \ac{MDG} and \ac{PDG} are different. 
The \ac{MDG} focuses on distinct attention heads, while \ac{PDG} pushes each attention head to attend to different tokens/positions of the inputs.
So combining \ac{MDG} and \ac{PDG} generally leads to an improvement of BERT on almost all layers.
For example, the performances increase by 0.21\% and 1.51\%, after being guided by the \ac{MDG} and \ac{PDG} separately at layer 6, while the improvement reaches 1.76\% after combining \ac{MDG} and \ac{PDG}.
Moreover, lower layers were found to perform broad attention across all pairs of tokens~\cite{DBLP:conf/blackboxnlp/ClarkKLM19}. 
Therefore, lower layers call for the guidance of \ac{PDG} to promote the diversity of tokens rather than \ac{MDG}, compared to the middle and top layers.

%Middle layers were found to mostly capture transferable syntactic and semantic knowledge~\cite{DBLP:conf/naacl/HewittM19,DBLP:conf/acl/TenneyDP19}.
%It is not surprising that the upper layers do not perform well, as these layers are specifically tuned towards the pre-training tasks of BERT – masked language modeling and next-sentence prediction – not the classification on the MedNLI dataset.

\subsection{Effect of \ac{AG} on different training sizes}
\label{sec:train_size}
\begin{figure}[h]
\centering  
\includegraphics[width=0.4\textwidth]{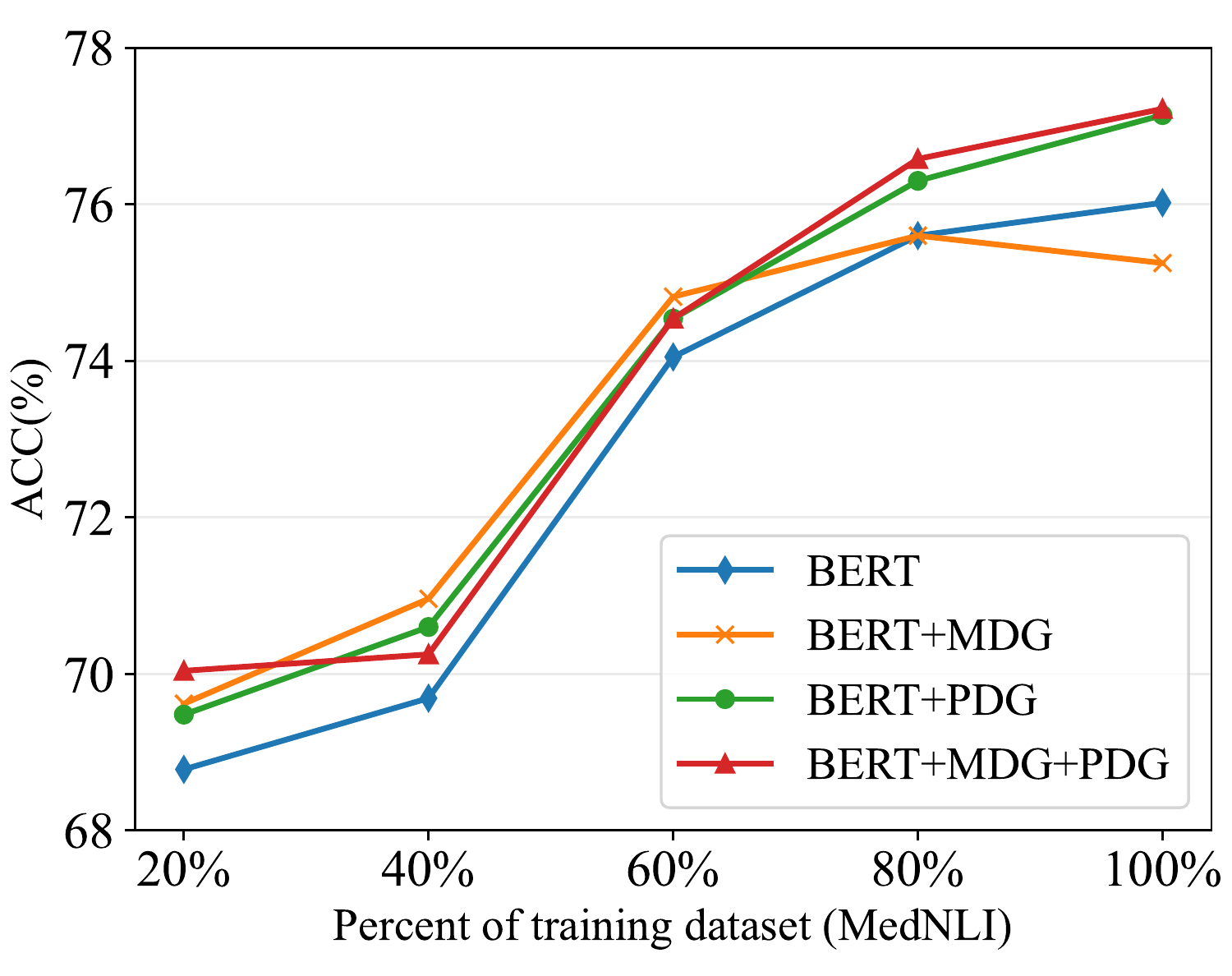}  
\caption{Performance of BERT with \ac{AG} on different amounts of training data.} 
\label{trainsize}
\end{figure}

In Table \ref{result}, we found that the proposed \ac{AG} brings more improvements on MedNLI and Cross-Genre-IR datasets than on the MultiNLI dataset.
To explore whether our \ac{AG} mechanism is affected by the size of the training datasets, we randomly select 20\% to 100\% data from the training set of MedNLI for fine-tuning. 
The detailed results are illustrated in Figure~\ref{trainsize}.
% We confirm the following two conclusions after this experiments:
We have the following observations in this experiment.

The \ac{PDG} and \ac{MDG} can improve BERT at different training sizes generally, even though the data size is small.
Specifically, when only 20\% of the training set is used, the \ac{MDG} increases the Accuracy by 0.84\% (i.e., from 68.78\% to 69.62\%) while \ac{PDG} also improves the Accuracy by 0.70\% (from 68.78\% to 69.48\%). 
The reasons of \ac{MDG} and \ac{PDG} are effective on small datasets are that when there is not enough training data, it is difficult for \acp{PLM} to be adapted to a different task.
In other words, the self-attention is not well fine-tuned. Thus, the guidance of self-attention becomes particularly important.
Moreover, it is easy for \acp{PLM} to over-fit on small datasets.
\ac{PDG} and \ac{PDG} could help in some ways to alleviate such over-fitting issues.

\subsection{Case study with attention heatmap}
We visualize the changes of attention heads with/without the guidance of \ac{MDG} and \ac{PDG} to see whether there is any diverse attention pattern after guidance.
We randomly select one sample from the test set of the MedNLI dataset and visualize the average attention map from all heads at all layers of BERT in Figure \ref{attention}.
To make it easy for observation, Figure \ref{attention} only shows the first 15 tokens in the sequence, and the selected token sequence is ``[ `[CLS]', `the', `patient', `denied', `any', `headache ', `,', `visual', `changes', `,', `chest', `pain', `,', `pl', `\#\#eur'] ".
% To facilitate comparison, we use blue boxes to mark the same area of different attention maps. 

As shown in Figure~\ref{attention}(d), the attention map with both \ac{MDG} and \ac{PDG} could pay attention to more positions compared with Figure~\ref{attention}(a) which is not guided by \ac{AG}.
For example, more attention is paid to the token in the last column (i.e., `\#\#eur'), which is overlooked by the attention map without \ac{AG} in Figure \ref{attention}(a).
In fact, the token `\#\#eur' and the previous token `pl' constitute an important medical concept `pleur', which should be paid more attention.
\ac{AG} can make such kinds of tokens get more attention, which is why \acp{PLM} can be improved by \ac{AG}.

% (通过怎样的比较得出怎样的结论，这段不是很清晰。而且有些结论并没有加分，反而可能会减分)

\subsection{Case study with attention principal component analysis}
In order to explore whether our \ac{AG} mechanism promotes the richness and diversity of self-attention, we randomly select some samples from the test set of MedNLI dataset and perform dimensionality reduction through \ac{PCA}~\cite{jolliffe2016principal} on BERT’s all attention heads from all layers. 
Figure \ref{pca} shows the spatial distribution of each attention head with/without the \ac{AG} mechanism. 
From Figure \ref{pca}, we can see that the attention distributions with \ac{MDG}, \ac{PDG} and both of them ((b)-(d) of Figure~\ref{pca}) are more dispersed than the distribution without \ac{AG} (see Figure~\ref{pca}(a)).
This suggests that the proposed \ac{AG} mechanism (\ac{MDG} and \ac{PDG} included) is effective, and \ac{AG} does encourage self-attention to pay attention to wider positions of the sequence.

Moreover, the distribution of multi-headed attention in Figure~\ref{pca}(b) (i.e., being guided by \ac{MDG}) is more scattered than in Figure~\ref{pca}(c) (i.e., being guided by \ac{PDG}).  
Obviously, the reason is that  \ac{MDG} is designed to push the diversity of different attention maps which will lead to scattered attention heads. 

\subsection{Time cost analysis}
Most of the previous studies directly modify the computation process of self-attention, e.g. \cite{DBLP:journals/corr/abs-2012-14642, DBLP:conf/www/XiaWTC21}, which means that they need to re-train the \acp{PLM}.
On the contrast, our method works in the fine-tuning phrase, and does not need to re-train the \acp{PLM}.
Thus, our \ac{AG} also has merits in terms of time cost.

Nevertheless, the calculation process of \ac{AG} will take more time than directly fine-tuning the pre-trained models on specific datasets.
% However, the extra time of \ac{AG} doesn't cost much.
Table \ref{time} shows the per-epoch training time of different \acp{PLM} with or without our \ac{AG} on three datasets. 

As can be seen from Table \ref{time},  the increased time cost is minor by adding \ac{AG}.
Specifically, the extra time cost by \ac{AG} per-epoch training is about 130 seconds, 7 seconds and 19 seconds on MultiNLI, MedNLI and Cross-genre-IR datasets, respectively. 
We consider the time cost acceptable, since \ac{AG} can improve different pre-trained models significantly.

\begin{table}[]
\centering
\caption{Per-epoch training time (in seconds) on the three datasets using different \acp{PLM} with or without the proposed \ac{AG}. 
}
%Significant improvements over the best baseline results are marked with $^\ast$ (t-test, $p < 0.05$).}
\begin{threeparttable}
% \resizebox{0.9\textwidth}{!}{
\begin{tabular}{l | c| c| c}
\toprule
 & MultiNLI & MedNLI & Cross-genre-IR\\
\midrule
BERT & 2190 & 95 & 339\\ 
BERT+AG & 2334 & 112 & 361\\ 			
\midrule
ALBERT & 2639 & 96 & 401 \\
ALBERT+AG & 2770 & 101 & 423\\
\midrule
Roberta & 2190  & 82 & 340\\
Roberta+AG & 2344 & 87 & 354\\		

\midrule
BioBERT& 2225 & 82 & 341\\
BioBERT+AG & 2318 & 87 & 361\\
\midrule 
ClinicalBERT & 2228 & 82 & 339 \\  
ClinicalBERT+AG & 2372 & 85 & 355\\			
\midrule 
BlueBERT   & 2210 & 100 & 340\\ 
BlueBERT+AG & 2324 & 102 & 359\\ 	
\midrule
SciBERT & 2215 & 81 & 338\\ 
SciBERT+AG & 2345 & 86 & 360\\			
\bottomrule       
\end{tabular}
\end{threeparttable}  
\label{time} 
\end{table}

\section{Related Work}
Existing studies on self-attention can be roughly classified into three groups: self-attention probing, self-attention revising and self-attention guiding.

\subsection{Self-attention probing}
This line of research focuses on the pattern probing of self-attention, i.e., the analysis of the interpretability of the weights and connections in particular.
For example, \citet{DBLP:conf/blackboxnlp/VigB19} visualize attentions and analyze the interaction between attention and syntax over a large corpus.
They find that different attentions target different parts of speech at different layers of the model, and that the attentions align with the dependency relations closely, especially in the middle layers.
Similarly, \citet{DBLP:conf/blackboxnlp/ClarkKLM19} demonstrate through visual analysis and statistical analysis that the substantial syntactic information is captured in BERT’s attentions.
\citet{DBLP:conf/emnlp/KovalevaRRR19} summarize 5 kinds of frequent attention patterns, called vertical, diagonal, vertical+diagonal, block, and heterogeneous, respectively.
\citet{DBLP:conf/acl/VoitaTMST19} identify the most important heads in each encoder layer using layer-wise relevance propagation, and then attempt to characterize the roles they perform.
\citet{DBLP:journals/corr/abs-2103-14625} present DODRIO, an open-source interactive visualization tool to help researchers and practitioners analyze attention mechanisms to meet the needs of self-attention visualization. 

\subsection{Self-attention revising}
This line of research modifies the attention formula to bias the attention weights towards local areas~\cite{DBLP:conf/acl/XuWYZC19,DBLP:journals/corr/abs-1905-06596,DBLP:journalsYangLKXXSXY20}.
For example, \citet{DBLP:conf/iclr/WuFBDA19} and \citet{DBLP:conf/naacl/YangWWCT19} use convolutional modules to replace self-attentions in some parts, making the networks computationally more efficient.
\citet{DBLP:conf/emnlp/RaganatoST20} design seven predefined patterns, each of which takes the place of an attention head to train Neural Machine Translation models without the need of learning them. 
The advantage of this method is that it can reduce the parameter footprint without loss of translation quality.
\citet{DBLP:journals/corr/abs-2012-14642} consider direction mask, word distance mask, and dependency distance mask simultaneously, and add them into the attention calculation to obtain the structural priors.
Similarly, \citet{DBLP:conf/acl/LiZLXC21} map each token into a tree node, and calculate the distance of any two nodes, after which the distance is added to the attention calculation.
\citet{DBLP:conf/www/XiaWTC21} inject word similarity knowledge into the attention calculation to make the BERT model aware of the word pair similarity.
\vspace{-2mm}
\subsection{Self-attention guiding}
Different from the above two research lines, our work belongs to self-attention guiding, which guides the learning of self-attention without introducing any new parameters or modifying the attention calculation formulas.
\citet{DBLP:conf/emnlp/DeshpandeN20}'s work belongs to this category.
In their work, five fixed patterns are predefined based on the analyses of attentions, based on which a regularization term is added to force the attentions to approach the predefined attention patterns in the training phase of \acp{PLM}.
There are at least two differences compared with our work.
First, we do not need to predefine attention patterns.
Instead, the attention is guided adaptively through the \ac{MDG} and \ac{PDG} parts.
Second, we do not need to train \acp{PLM} from scratch.
Our attention guiding method works in the fine-tuning phase of the \acp{PLM}.
We compared with \citet{DBLP:conf/emnlp/DeshpandeN20}'s work in Table~\ref{knowledge}, demonstrating that our method achieves comparable or better performance without introducing new knowledge or predefining attention patterns beforehand.
\vspace{-2mm}

\section{Conclusion and Future Work}
In this work, we have proposed two kinds of attention guiding methods, i.e., the \acf{MDG} and the \acf{PDG}, to improve the performance of \acp{PLM} by encouraging the learned attentions to derive more information from the inputs and to be more diverse.
Experimental results of seven \acp{PLM} on three datasets have validated the effectiveness of our proposed methods. 
Especially, we have found that the proposed attention guiding mechanism works on small datasets and large datasets, which is attractive as building large labeled dataset is time consuming and labor intensive.

As to future work, we plan to explore how to incorporate more domain-specific knowledge to guide self-attention learning in low-resource domains, e.g., the relations of diseases, drugs, and symptoms in medical domains.

\section*{Reproducibility}
This work uses publicly available data.
To facilitate the reproducibility of the reported results, we release the code at \url{https://anonymous.4open.science/r/attentionGuiding-F6C0}.

% \begin{acks}
% All content represents the opinion of the authors, which is not necessarily shared or endorsed by their respective employers and/or sponsors.
% \end{acks}
%\clearpage

\bibliographystyle{ACM-Reference-Format}
\bibliography{references}

\end{document}